%% file: main.tex
\documentclass[sigconf]{acmart}

\renewcommand\footnotetextcopyrightpermission[1]{} 
\AtBeginDocument{%
  \providecommand\BibTeX{{%
    \normalfont B\kern-0.5em{\scshape i\kern-0.25em b}\kern-0.8em\TeX}}}

\usepackage{balance}
\usepackage[ruled,vlined]{algorithm2e}
\usepackage{bbding}
\usepackage{enumitem}

\usepackage{array}
\newcolumntype{L}[1]{>{\raggedright\let\newline\\\arraybackslash\hspace{0pt}}m{#1}}
\newcolumntype{C}[1]{>{\centering\let\newline\\\arraybackslash\hspace{0pt}}m{#1}}
\newcolumntype{R}[1]{>{\raggedleft\let\newline\\\arraybackslash\hspace{0pt}}m{#1}}
\usepackage{delarray}

\usepackage{xparse}

\ExplSyntaxOn
\NewDocumentCommand{\setarray}{mm}{\clist_clear_new:c { l_jens_#1_array_clist} \clist_set:cn {l_jens_#1_array_clist}{#2}}
 
\NewExpandableDocumentCommand{\listarray}{mm}{\clist_item:cn { l_jens_#1_array_clist}{#2}}
\ExplSyntaxOff

\usepackage[normalem]{ulem}

\usepackage{tikz}
\usetikzlibrary{arrows, arrows.meta,tikzmark,shadows,calc,shapes,snakes}

\usepackage{color}

\newcommand{\nop}[1]{}

\newcommand\WORK{\textsc{Auto-Split}}
\newcommand\EDGEONLY{\textsc{Edge-Only}}
\newcommand\CLOUDONLY{\textsc{Cloud-Only}}
\newcommand\SPLIT{\textsc{SPLIT}}
\newtheorem{remark}{Remark}

\newcommand*\whitecircle[1]{\tikz[baseline=(char.base)]{  \node[shape=circle,fill=white,draw,inner sep=1.5pt] (char) {\color{black}#1};}}

\newcommand{\st}{\mbox{s.t.}\hspace{1pt}}

\usepackage{mathtools,bbm}
\usepackage{hyperref}

\begin{document}

\title[\textsc{Auto-Split}: A General Framework of Collaborative Edge-Cloud AI]{\textsc{Auto-Split}: A General Framework\\ of Collaborative Edge-Cloud AI}

\author{Amin Banitalebi-Dehkordi}
\authornote{Both authors contributed equally. Correspondence: amin.banitalebi@huawei.com}
\author{Naveen Vedula}
\authornotemark[1]
\affiliation{%
  \institution{Huawei Technologies Canada Co. Ltd.}
  \city{Vancouver}
  \country{Canada}
}

\author{Jian Pei}
\affiliation{%
  \institution{School of Computing Science\\ Simon Fraser University}
  \city{Vancouver}
  \country{Canada}
}

\author{Fei Xia}
\affiliation{%
  \institution{Huawei Technologies}
  \city{Shenzhen}
  \country{China}
}

\author{Lanjun Wang}
\author{Yong Zhang}
\affiliation{%
  \institution{Huawei Technologies Canada Co. Ltd.}
  \city{Vancouver}
  \country{Canada}
}

\renewcommand{\shortauthors}{Banitalebi-Dehkordi and Vedula, et al.}

\fancyhead{}


\input{00_abstract.tex}



\maketitle


\input{01_intro.tex}

\input{02_background.tex}

\input{03_motivation.tex}

\input{04_formulation.tex}

\input{04_auto_split.tex}
\input{05_experiments.tex}

\input{06_conclusion.tex}

\bibliographystyle{ACM-Reference-Format}
\bibliography{references}

\clearpage
\input{07_section_supplementary.tex}

\end{document}

%% file: 00_abstract.tex

\begin{abstract}

In many industry scale applications, large and resource consuming machine learning models reside in powerful cloud servers.  At the same time, large amounts of input data are collected at the edge of cloud. The inference results are also communicated to users or passed to downstream tasks at the edge.  The edge often consists of a large number of low-power devices. It is a big challenge to design industry products to support sophisticated deep model deployment and conduct model inference in an efficient manner so that the model accuracy remains high and the end-to-end latency is kept low. This paper describes the techniques and engineering practice behind \WORK{}, an edge-cloud collaborative prototype of Huawei Cloud. This patented technology is already validated on selected applications, is on its way for broader systematic edge-cloud application integration, and is being made available for public use as an automated pipeline service for end-to-end cloud-edge collaborative intelligence deployment. To the best of our knowledge, there is no existing industry product that provides the capability of Deep Neural Network (DNN) splitting. \footnote{Code and demo are available at: \url{https://marketplace.huaweicloud.com/markets/aihub/notebook/detail/?id=5fad1eb4-50b2-4ac9-bcb0-a1f744cf85c7}}

\nop{
   There is a significant demand for efficient execution of neural networks inference on edge devices. However, limited computation capability available at the low-power edge devices imposes a major challenge. Two existing solutions to alleviate this problem are neural network compression  and running models on the cloud, both with limitations and deficiencies. This paper proposes an alternative solution that addresses the shortcomings of these two methods by distributing the inference between the edge and cloud. In our method, a part of the network runs in lower mixed precision bit-widths at the edge, and the rest runs on the cloud. We propose to jointly solve the split identification and bit-width assignment. Our optimization formulation finds a balance in the trade-off among the model accuracy, edge device capacity, transmission cost, and the overall latency. Extensive experiments show that this method outperforms existing solutions in many practical scenarios.
}
\end{abstract}

%% file: 01_intro.tex
\section{Introduction}
\label{sec:introduction}

The recent exciting advances in AI are heavily driven by two spurs, large scale deep learning models and huge amounts of data.  The current AI flying wheel trains large scale deep learning models by harnessing large amounts of data, and then applies those models to tackle and harvest even larger amounts of data in more applications.

Large scale deep models are typically hosted in cloud servers with unrelenting computational power.  At the same time, data is often distributed at the edge of cloud, that is, the edge of various networks, such as smart-home cameras, authorization entry (e.g. license plate recognition camera), smart-phone and smart-watch AI applications, surveillance cameras, AI medical devices (e.g. hearing aids, and fitbits), and IoT nodes. The combination of powerful models and rich data achieve the wonderful progress of AI applications.

However, the gap between huge amounts of data and large deep learning models remains and becomes a more and more arduous challenge for more extensive AI applications. Connecting data at the edge with deep learning models at cloud servers is far from straightforward.  Through low-power devices at the edge, data is often collected, and machine learning results are often communicated to users or passed to downstream tasks. Large deep learning models cannot be loaded into those low-power devices due to the very limited computation capability.
Indeed, deep learning models are becoming more and more powerful and larger and larger. The grand challenge for utilization of the latest extremely large models, such as GPT-3 \cite{brown2020language} (350GB memory and 175B parameters in case of GPT-3), is far beyond the capacity of just those low-power devices. For those models, inference is currently conducted on cloud clusters. It is impractical to run such models only at the edge.




Uploading data from the edge devices to cloud servers is not always desirable or even feasible for industry applications.  Input data to AI applications is often generated at the edge devices. It is therefore preferable to execute AI applications on edge devices (the \EDGEONLY{} solution).
Transmitting high resolution, high volume input data all to cloud servers (the \CLOUDONLY{} solution) may incur high transmission costs, and may result in high end-to-end latency. Moreover, when original data is transmitted to the cloud, additional privacy risks may be imposed.

\begin{figure}[t]
    \centering
    \includegraphics[width=1.0\linewidth]{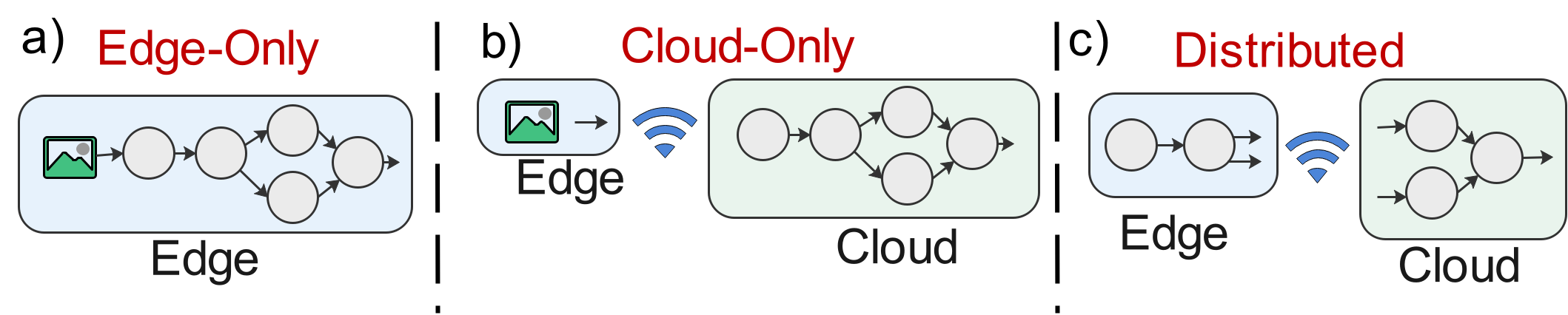}
    \caption{Different approaches of edge-cloud collaboration}
    \label{fig:various_approaches}
\end{figure}


In general, there are two types of existing industry solutions to connect large models and large amounts of data: \EDGEONLY{} and Distributed approaches (See Fig. \ref{fig:various_approaches}). The \EDGEONLY{} solution applies model compression to force-fit an entire AI application on edge devices.  This approach may suffer from serious accuracy loss~\cite{cheng2017survey}.  

Alternatively, one may follow a distributed approach to execute the model partially on the edge and partially on the cloud.  The distributed approach can be further categorized into three subgroups.  First, the \CLOUDONLY{} approach conducts inference on the cloud. It may incur high data transmission costs, especially in the case of high resolution input data for high-accuracy applications. Second, the cascaded edge-cloud inference approach divides a task into multiple sub-tasks, deploys some sub-tasks on the edge and transmits the output of those tasks to the cloud where the other tasks are run. Last, a multi-exit solution deploys a lightweight model on the edge, which processes the simpler cases, and transmits the more difficult cases to a larger model in the cloud servers. The cascaded edge-cloud inference approach and the multi-exit solution are application specific, and thus are not flexible for many use-cases. Multi-exit models may also suffer from low accuracy and have non-deterministic latency.   A series of products and services across the industry by different cloud providers \cite{AmazonNeo} are developed, such as SageMaker Neo, PocketFlow, Distiller, PaddlePaddle, etc.


Very recently, an edge-cloud collaborative approach has been explored, mainly from academia~\cite{neurosurgeon, dads, QDMP}.  The approach exploits the fact that the data size at some intermediate layer of a deep neural network (DNN for short) is significantly smaller than that of raw input data. This approach partitions a DNN graph into edge DNN and cloud DNN, thereby reduces the transmission cost and lowers the end-to-end latency~\cite{neurosurgeon, dads, QDMP}. The edge-cloud collaborative approach is generic, can be applied to a large number of AI applications, and thus represents a promising direction for industry. To foster industry products for the edge-cloud collaborative approach, the main challenge is to develop a general framework to partition deep neural networks between the edge and the cloud so that the end-to-end latency can be minimized and the model size in the edge can be kept small. Moreover, the framework should be general and flexible so that it can be applied to many different tasks and models.

In this paper, we describe the techniques and engineering practice behind \WORK{}, an edge-cloud collaborative prototype of Huawei Cloud. This patented technology is already validated on selected applications, such as license plate recognition systems with HiLens edge devices \cite{HiLens, ModelArtsPro-Deployment, SDC}, is on its way for broader systematic edge-cloud application integration \cite{MindXEdge}, and is being made available for public use as an automated pipeline service for end-to-end cloud-edge collaborative intelligence deployment \cite{ModelArtsEdge}. To the best of our knowledge, there is no existing industry product that provides the capability of DNN splitting.


Building an industry product of DNN splitting is far from trivial, though we can be inspired by some initial ideas partially from academia.  The existing edge-cloud splitting techniques reported in literature still incur substantial end-to-end latency.  One innovative idea in our design is the integration of edge-cloud splitting and post-training quantization.  We show that by jointly applying edge-cloud splitting along with post-training quantization to the edge DNN, the transmission costs and model sizes can be reduced further, and results in reducing the end-to-end latency by 20--80\% compared to the current state of the art, QDMP \cite{QDMP}.

\begin{figure*}
    \centering
    \includegraphics[width=0.9\linewidth]{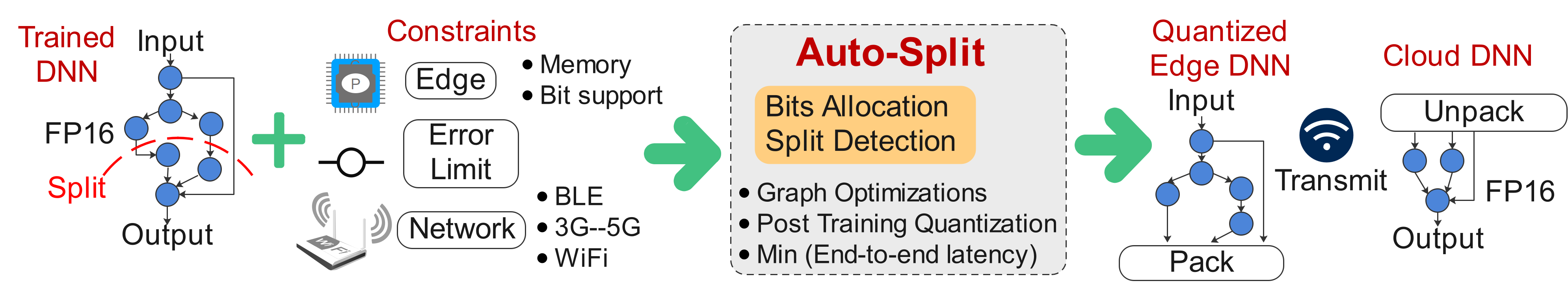}
    \caption{\WORK{} overview: inputs are a trained DNN and the constraints, and outputs are optimal split and bit-widths.}
    \label{fig:auto_split_overview}
\end{figure*}

Fig. \ref{fig:auto_split_overview} shows an overview of our framework. We take a trained DNN with sample profiling data as input, and apply several environment constraints to optimize for end-to-end latency of the AI application. \WORK{} partitions the DNN into an edge DNN to be executed on the edge device and a cloud DNN for the cloud device. \WORK{} also applies post-training quantization on the edge DNN and assigns bit-widths to the edge DNN layers (works offline). \WORK{} considers the following constraints: a) edge device constraints, such as on-chip and off-chip memory, number and size of NN accelerator engines, device bandwidth, and bit-width support,  b) network constraints, such as uplink bandwidth based on the network type (e.g., BLE, 3G, 5G, or WiFi), c) cloud device constraints, such as memory, bandwidth and compute capability, and d) required accuracy threshold provided by the user.

The rest of the paper is organized as follows. We discuss related works in Section \ref{sec:background}. Sections \ref{sec:formulation} and \ref{sec:solution} cover our design, formulation, and the proposed solution. 
In Section \ref{sec:experiment}, we present a systematical empirical evaluation, and a real-world use-case study of \WORK{} for the task of license plate recognition. 
The Appendix consists of further implementation details and ablation studies.

%% file: 02_background.tex
\section{Related Works}
\label{sec:background}
Our work is related to edge cloud partitioning and model compression using mixed precision post-training quantization. We focus on distributed inference and assume that models are already trained. 

\subsection{Model Compression (On-Device Inference)}

Quantization methods \cite{dorefa,pact,hawq} compress a model by reducing the bit precision used to represent parameters and/or activations. Quantization-aware training performs bit assignment during training \cite{zhang2018lq, esser2019learned, dorefa}, while post-training quantization applies on already trained models \cite{Jacob2018QuantizationAT,Zhao2019ImprovingNN, banner2018aciq, Krishnamoorthi2018QuantizingDC}. Quantizing all layers to a similar bit width leads to a lower compression ratio, since not all layers of a DNN are equally sensitive to quantization. Mixed precision post-training quantization is proposed to address this issue \cite{zeroq, hawq, dnas, haq}.

There are other methods besides quantization for model compression \cite{hinton2015distilling, gao2020rethinking, polino2018model, mishra2017apprentice, cheng2017survey}.  These methods have been proposed to address the prohibitive memory footprint and inference latency of modern DNNs. They are typically orthogonal to quantization and  include techniques such as distillation ~\cite{hinton2015distilling}, pruning ~\cite{cheng2017survey, he2017channel, gao2020rethinking}, or combination of pruning, distillation and quantization  ~\cite{polino2018model, cheng2017survey, han2015deep}.

\subsection{Edge Cloud Partitioning}
\label{sec:background_distributed}




Related to our work are progressive-inference/multi-exit models, which are essentially networks with more than one exit (output node). A light weight model is stored on the edge device and returns the result as long as the minimum accuracy threshold is met. Otherwise, intermediate features are transmitted to the cloud to execute a larger model. A growing body of work from both the research \cite{teerapittayanon2016branchynet, spinn, MultiScale, SCANAS, EdgeAI} and industry ~\cite{nervana_early_exit, tenstorrent} has proposed transforming a given model into a progressive inference network by introducing intermediate exits throughout its depth. So far the existing works have mainly explored hand-crafted techniques and are applicable to only a limited range of applications, and the latency of the result is non-deterministic (may route to different exits depending on the input data) \cite{zhou2017adaptive, teerapittayanon2016branchynet}. Moreover, they require retraining, and re-designing DNNs to implement multiple exit points \cite{teerapittayanon2016branchynet, spinn, SCANAS}.

The most closely related works to \WORK{}  are graph based DNN splitting techniques \cite{neurosurgeon, dads, QDMP, wang2018not}. These methods split a DNN graph into edge and cloud parts to process them in edge and cloud devices separately. They are motivated by the facts that: a) data size of some intermediate DNN layer is signiﬁcantly smaller than that of raw input data, and b) transmission latency is often a bottleneck in end-to-end latency of an AI application. Therefore, in these methods, the output activations of the edge device are significantly smaller in size, compared to the DNN input data.

\begin{figure}
    \centering
    \includegraphics[width=1.0\linewidth]{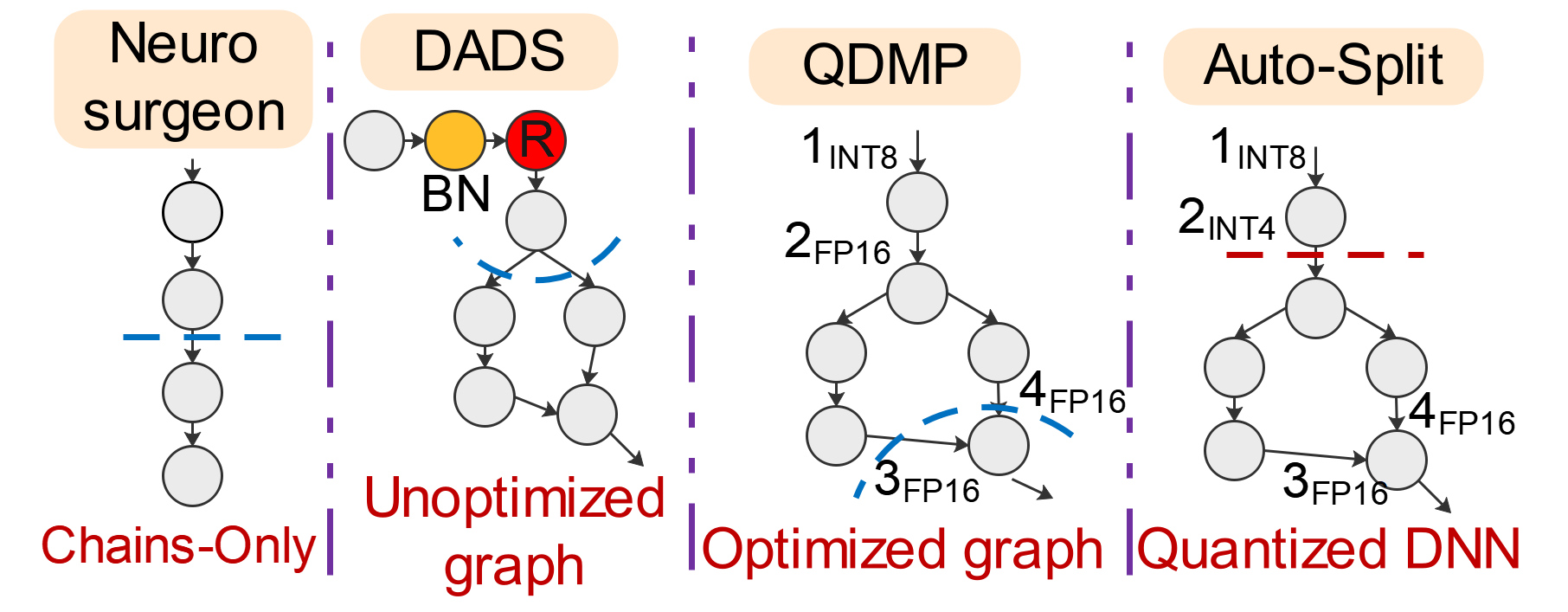}
    \caption{Different methods of edge-cloud partitioning. Neurosurgeon~\cite{neurosurgeon}:  handles chains only. DADS ~\cite{dads}: min-cut on un-optimized DNN. QDMP ~\cite{QDMP}: min-cut on optimized DNN. All three use Float models. \WORK{} explores new search space with joint mixed-precision quantization of edge and split point identification. $BN:$ Batch norm, $R:$ Relu. See \S \ref{sec:background_distributed}}
        
    \label{fig:related_work}
\end{figure}

Fig. \ref{fig:related_work} shows a comparison between the existing works and \WORK{}.
Earlier techniques such as Neurosurgeon \cite{neurosurgeon} look at primitive DNNs with chains of layers stacked one after another and cannot handle state-of-the-art DNNs which are implemented as complex directed acyclic graphs (DAGs). Other works such as DADS \cite{dads} and QDMP~\cite{QDMP} can handle DAGs, but only for floating point DNN models. These work requires two copies of entire DNN, one stored on the edge device and the other on the cloud, so that they can dynamically partition the DNN graph depending on the network speed. Both DADS and QDMP assume that the DNN fits on the edge device which may not be true especially with the growing demand of adding more and more AI tasks to the edge device.
It is also worth noting that methods such as DADS do not consider inference graph optimizations such as batchnorm folding and activation fusions. These methods apply the min-cut algorithm on an un-optimized graph, which results in an sub-optimal split \cite{QDMP}.

The existing works on edge cloud splitting, do not consider quantization for split layer identification. Quantization of the edge part of a DNN enlarges the search space of split solutions. For example, as shown in Fig. ~\ref{fig:related_work}, high transmission cost of a float model at node 2, makes this node a bad split choice. However, when quantized to 4-bits, the transmission cost becomes lowest compared to other nodes and this makes it the new optimal split point.

\subsection{Related Services and Products In Industry}
The existing services and products in the industry are mostly made available by cloud providers to their customers. Examples include Amazon SageMaker Neo \cite{AmazonNeo}, Alibaba Astraea, CDN, ENS, and LNS \cite{fu2020astraea, AliCDN, AliAPI}, Tencent PocketFlow \cite{PocketFlow}, Baidu PaddlePaddle \cite{PaddlePaddle}, Google Anthos \cite{GoogleAnthos}, Intel Distiller \cite{distiller}.
Some of these products started by targeting model compression services, but are now pursuing collaborative directions.
It is also worth noting that cloud providers generally seek to build ecosystems to attract and keep their customers. Therefore, they may prefer the edge-cloud collaborative approaches over the edge-only solutions. This will also give them better protection and control over their IP.

%% file: 03_motivation.tex



%% file: 04_formulation.tex
\section{Problem Formulation} \label{sec:formulation}
In this section, we first introduce some basic setup. Then, we formulate a nonlinear integer optimization problem for \WORK{}. This problem minimizes the overall latency under an edge device memory constraint and a user given error constraint, by jointly optimizing the split point and bid-widths for weights and activation of layers on the edge device. 


\subsection{Basic Setup}
Consider a DNN with $N\in \mathbb{Z}$ layers, where $\mathbb{Z}$ defines a non-negative integer set. Let $\textbf{s}^w \in \mathbb{Z}^N$ and $\textbf{s}^a\in \mathbb{Z}^N$ be vectors of sizes for weights and activation, respectively. Then, $\textbf{s}^w_i$ and $\textbf{s}^a_i$ represent sizes for weights and activation at the $i$-th layer, respectively. For the given DNN, both $\textbf{s}^w$ and $\textbf{s}^a$ are fixed. Next, let $\textbf{b}^w \in \mathbb{Z}^N$ and $\textbf{b}^a\in \mathbb{Z}^N$ be bit-widths vectors for weights and activation, respectively. Then, $\textbf{b}^w_i$ and $\textbf{b}^a_i$ represent bit-widths for weights and activation at the $i$-th layer, respectively.

Let $L^{edge}(\cdot)$ and $L^{cloud}(\cdot)$ be latency functions for given edge and cloud devices, respectively. For a given DNN, $\textbf{s}^w $ and $\textbf{s}^a$ are fixed. Thus, $L^{edge}$ and $L^{cloud}$ are functions of weights and activation bit-widths. We denote latency of executing the $i$-th layer of the DNN on edge and cloud by $ \mathcal{L}^{edge}_{i} = L^{edge}(\textbf{b}^w_i, \textbf{b}^a_i)$ and $ \mathcal{L}^{cloud}_{i} = L^{cloud}(\textbf{b}^w_i, \textbf{b}^a_i)$, respectively. We define a function $L^{tr}(\cdot)$ which measures latency for transmitting data from edge to cloud, and then denote the transmission latency for the $i$-th layer by  $ \mathcal{L}^{tr}_{i} = L^{tr}(\textbf{s}^a_i \times \textbf{b}^a_i)$.

To measure quantization errors, we first denote  $w_i(\cdot)$ and $a_i(\cdot)$ as weights and activation vectors for a given bit-width at the $i$-th layer. Without loss of generality , we assume the bit-widths for the original given DNN are $16$. Then by using the mean square error function $MSE(\cdot, \cdot)$, we denote the quantization errors at the $i$-th layer for weights and activation by  $\mathcal{D}^w_i =  MSE\big( w_i(16), w_i(\textbf{b}^w_i) \big)$ and $\mathcal{D}^a_i =  MSE\big( a_i(16), a_i(\textbf{b}^a_i) \big)$, respectively. Notice that MSE is a common measure for quantization error and has been widely used in related studies such as \cite{banner2019post, banner2018aciq, Zhao2019ImprovingNN, choukroun2019low}. It is also worth noting that while we follow the existing literature on using the MSE metric, other distance metrics such as cross-entropy or KL-Divergence can alternatively be utilized without changing our algorithm.

\subsection{Problem Formulation}
In this subsection, we first describe how to write \WORK{}'s objective function in terms of latency functions and then discuss how we formulate the edge memory constraint and the user given error constraint. We conclude this subsection by providing a nonlinear integer optimization problem formulation for \WORK{}.

\paragraph{\textbf{Objective function:}} Suppose the DNN is split at layer $n\in\{z\in\mathbb{Z}|0\leq z\leq N\}$ , then we can define the objective function by summing all the latency parts, i.e.,
\begin{equation}
\label{eq:latency_proposed}
\begin{split}
    \mathcal{L}(\textbf{b}^w, \textbf{b}^a, n) = \sum_{i=1}^{n} \mathcal{L}^{edge}_{i} + \mathcal{L}^{tr}_{n} +
    \sum_{i=n+1}^{N} \mathcal{L}^{cloud}_{i}.
\end{split}
\end{equation}
When $n=0$ (resp. $n=N$), all layers of the DNN are executed on cloud (resp., edge). Since cloud does not have resource limitations (in comparison to the edge), we assume that the original bit-widths are used to avoid any quantization error when the DNN is executed on the cloud. Thus,  $\mathcal{L}^{cloud}_{i}$ for $i=1,\ldots,N$ are constants. Moreover, since $\mathcal{L}^{tr}_{0}$ represents the time cost for transmitting raw input to the cloud, it is reasonable to assume that $\mathcal{L}^{tr}_{0}$ is a constant under a given network condition. Therefore, the objective function for \CLOUDONLY{} solution $\mathcal{L}(\textbf{b}^w, \textbf{b}^a, 0)$ is also a constant. 

To minimize $\mathcal{L}(\textbf{b}^w, \textbf{b}^a, n)$, we can equivalently minimize 
\[
\begin{split}
& \mathcal{L}(\textbf{b}^w, \textbf{b}^a, n) -\mathcal{L}(\textbf{b}^w, \textbf{b}^a, 0) \\
&=\left(\sum_{i=1}^{n} \mathcal{L}^{edge}_{i} + \mathcal{L}^{tr}_{n} +
    \sum_{i=n+1}^{N} \mathcal{L}^{cloud}_{i}\right) - \left(\mathcal{L}^{tr}_{0} +
    \sum_{i=1}^{N} \mathcal{L}^{cloud}_{i}\right) \\
&= \left(\sum_{i=1}^{n} \mathcal{L}^{edge}_{i} + \mathcal{L}^{tr}_{n}
   \right) - \left(\mathcal{L}^{tr}_{0} +
    \sum_{i=1}^{n} \mathcal{L}^{cloud}_{i}\right)
\end{split}
\]
After removing the constant $\mathcal{L}^{tr}_{0}$, we write our objective function for the \WORK{} problem as 
\begin{equation}
\label{eq:latency_simplified}
    \sum_{i=1}^{n} \mathcal{L}^{edge}_{i} + \mathcal{L}_n^{tr} -  \sum_{i=1}^{n} \mathcal{L}^{cloud}_{i}. 
\end{equation}

\begin{figure*}
    \centering
    \includegraphics[width=0.90\linewidth]{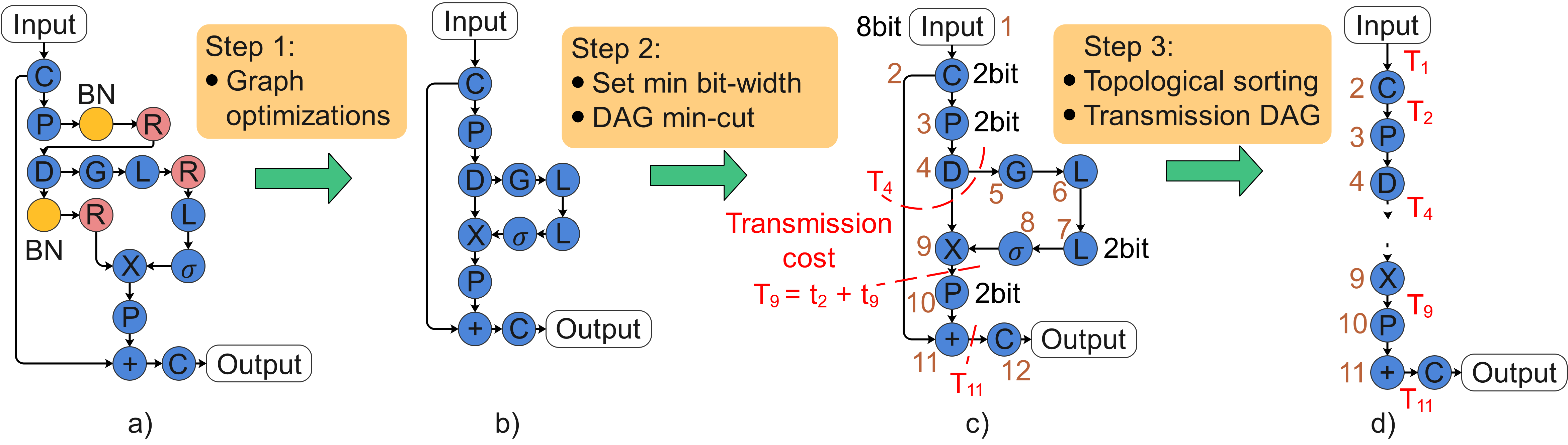}
    \caption{ An example of graph processing and the steps required to calculate the list of potential split points. 4a) is an example of inverted residual layer with squeeze \& excitation from MnasNet. Step1: DAG optimizations for inference such as batchnorm folding and activation fusion. Step2: Set $\textbf{b}_i^{a} = b_{min}$ for all activations, create weighted graph, and find the min-cut between sub graphs. Step 3: Apply topological sorting to create transmission DAG where nodes are layers and edges are transmission costs. $C:$Convolution $P:$Pointwise convolution, $D:$Depthwise convolution, $L:$Linear, $G:$Global pool, $BN:$Batch norm, $R:$Relu.}
    \label{fig:graph_optimization}
\end{figure*}

\paragraph{\textbf{Memory constraint:}} 
In hardware, ``read-only" memory stores the parameters (weights), and ``read-write" memory stores the activations~\cite{rusci2019memory}. The weight memory cost on the edge device can be
calculated with $\mathcal{M}^{w} = \sum_{i=1}^n (\textbf{s}_{i}^{w} \times \textbf{b}_{i}^{w})$. Unlike the weight memory, for activation memory only partial input activation and partial output activation need to be stored in the ``read-write" memory at a time. Thus, the memory required by activation is equal to the largest working set size of the activation layers at a time.
In case of a simple DNN chain, i.e., layers stacked one by one, the activation working set can be computed as $ \mathcal{M}^{a} = \underset{i=1,\ldots,n}{\max}{(\textbf{s}_{i}^{a} \times \textbf{b}_{i}^a)}$.
However, for complex DAGs, it can be calculated from the DAG. For example, in Fig. \ref{fig:graph_optimization}, when the depthwise layer is being processed, both the output activations of layer: 2 (convolution)  and layer: 3 (pointwise convolution) need to be kept in memory. Although the output activation of layer: 2 is not required for processing the depthwise layer, it needs to be stored for future layers such as layer 11 (skip connection). Assuming the available memory size of the edge device for executing the DNN is $M$, then the memory constraint for the \WORK{} problem can be written as 
\begin{equation}
\label{eq:memory}
   \mathcal{M}^{w} + \mathcal{M}^{a} \leq M. 
\end{equation}

\paragraph{\textbf{Error constraint:}} In order to maintain the accuracy of the DNN, we constrained the total quantization error by a user given error tolerance threshold $E$. In addition, since we assume that the original bit-widths are used for the layers executing on the cloud,
we only need to sum the quantization error on the edge device. Thus, we can write the error constraint for \WORK{} problem as
\begin{equation}
\label{eq:error}
    \sum_{i=1}^n  \left(\mathcal{D}^w_i + \mathcal{D}^a_i\right) \leq E. 
\end{equation}

\paragraph{\textbf{Formulation:}}
To summarize, the \WORK{} problem can be formulated based on the objective function \eqref{eq:latency_simplified} along with the memory  \eqref{eq:memory}  and error  \eqref{eq:error} constraints
\begin{subequations} \label{eq:optimization}
\begin{align}
    \underset{\textbf{b}^w,\textbf{b}^a \in \mathbb{B}^n, n}{\mathrm{min}} &\left( \sum_{i=1}^{n} \mathcal{L}_{i}^{edge} + \mathcal{L}_{n}^{tr}  - \sum_{i=1}^{n} \mathcal{L}_{i}^{cloud} \right) \label{eq:obj} \hspace{30pt} \\
    \st{} &  \mathcal{M}^{w} + \mathcal{M}^{a} \leq M, \hspace{30pt} \label{eq:con-1}\\
     & \sum_{i=1}^n  \left(\mathcal{D}^w_i + \mathcal{D}^a_i\right) \leq E, \label{eq:con-2}\hspace{31pt}
\end{align}
\end{subequations}
where $\mathbb{B}$ is the candidate bit-width set. Before ending this section, we make the following four remarks about problem \eqref{eq:optimization}.
\begin{remark}
 For a given edge device, the candidate bit-width set $\mathbb{B}$ is fixed. A typical $\mathbb{B}$ can 
 be $\mathbb{B} = \{2,4,6,8\}$ \cite{pulpnn}.
\end{remark}
\begin{remark}
The latency functions are not given explicitly \cite{netadapt,pnas,chamnet} in practice. Thus, simulators like \cite{samajdar2018scale, timeloop, accelergy} are commonly used to obtain latency \cite{choi2020prema, yang2018energy, zhe2020towards2bit}.
\end{remark}
\begin{remark}
 Since the latency functions are not explicitly defined  \cite{netadapt,pnas,chamnet} and the error functions are also nonlinear, problem \eqref{eq:optimization} is a nonlinear integer optimization function and NP-hard to solve. However, problem \eqref{eq:optimization} does have a feasible solution, i.e., $n=0$, which implies executing all layers of the DNN on cloud.
\end{remark}
\begin{remark}
\label{remark-4}
In practice, it is more tractable for users to provide accuracy drop tolerance threshold $A$, rather than the error tolerance threshold $E$. In addition, for a given $A$, calculating the corresponding $E$ is still intractable. To deal this issue, we tailor our algorithm in Subsection \ref{assignbit} to make it user friendly.
\end{remark}

%% file: 04_auto_split.tex
\section{\WORK{} Solution} \label{sec:solution}
As \WORK{} problem \eqref{eq:optimization} is NP-hard, we propose a multi-step search approach to find a list of potential solutions that satisfy \eqref{eq:con-1} and then select a solution which minimizes the latency and satisfies the error constraint \eqref{eq:con-2}. 

To find the list of potential solutions, we first collect a set of potential splits $\mathbb{P}$, by analyzing the data size that needs to be transmitted at each layer. Next, for each splitting point $n\in \mathbb{P}$, we solve two sets of optimization problems to generate a list of feasible solutions satisfying constraint \eqref{eq:con-1}. 

%
%
%
%
%

\subsection{Potential Split Identification} \label{subsection:split_identification}
As shown in Fig.~\ref{fig:graph_optimization}, to identify a list of potential splitting points, we preprocess the DNN graph by three steps.
First, we conduct graph optimizations  \cite{Jacob2018QuantizationAT} such as batch-norm folding and activation fusion on the original graph (Fig.~\ref{fig:graph_optimization}a) to obtain an optimized graph (Fig.~\ref{fig:graph_optimization}b).   
A potential splitting point $n$ should satisfy the memory limitation with lowest bit-width assignment for DNN, i.e., $b_{min}(\sum_{i=1}^n \textbf{s}_{i}^{w} + \underset{i=1,\ldots,n}{\max}{\textbf{s}_{i}^{a}) }\leq M$, where $b_{min}$ is the lowest bit-width constrained by the given edge device. Second, we create a weighted DAG as shown in Fig.~\ref{fig:graph_optimization}c, where nodes are layers and weights of edges are the lowest transmission costs, i.e., $b_{min}\textbf{s}^a$.  Finally, the weighted DAG is sorted in topological order and a new transmission DAG is created as shown in Fig.~\ref{fig:graph_optimization}d. Assuming the raw data transmission cost is a constant $T_0$, then a potential split point $n$ should have transmission cost $T_n \leq T_0$ (i.e., $\mathcal{L}^{tr}_n \leq \mathcal{L}^{tr}_0$). Otherwise, transmitting raw data to the cloud and executing DNN on the cloud is a better solution. Thus, the list of potential splitting points is given by
\begin{equation}
\label{eq:sort_solver_outputs}
    \mathbb{P} = \bigg\{n \in 0,1, \ldots, N|  T_n \leq T_0, \hspace{2pt} b_{min}(\sum_{i=1}^n \textbf{s}_{i}^{w} + \underset{i=1,\ldots,n}{\max}{\textbf{s}_{i}^{a}) }\leq M \bigg\}.
\end{equation}
\subsection{Bit-Width Assignment}
\label{assignbit}
In this subsection, for each $n\in \mathbb{P}$, we explore all feasible solutions which satisfy the constraints of \eqref{eq:optimization}.
As discussed in Remark \ref{remark-4}, explicitly setting $E$ is intractable. Thus, to obtain feasible solutions of \eqref{eq:optimization}, we first solve \begin{subequations} \label{eq:bit_allocation_opt-0}
\begin{align}
    \underset{\textbf{b}^w,\textbf{b}^a \in \mathbb{B}^n}{\mathrm{min}} & \sum_{i=1}^n  \left(\mathcal{D}^w_i + \mathcal{D}^a_i\right)  \hspace{35pt} \\
    \st{} 
    	      & \mathcal{M}^{w} +\mathcal{M}^{a} \leq M \label{budget},
\end{align}
\end{subequations}
and then select the solutions which are below the accuracy drop threshold $A$. 
We observe that for a given splitting point $n$, the search space of \eqref{eq:bit_allocation_opt-0} is exponential, i.e., $|\mathbb{B}|^{2n}$. To reduce the search space, we  
decouple problem \eqref{eq:bit_allocation_opt-0} into the following two problems
\begin{equation} \label{eq:bit_allocation_opt-1}
\begin{aligned}
    \underset{\textbf{b}^w \in \mathbb{B}^n}{\mathrm{min}} & \sum_{i=1}^n  \mathcal{D}^w_i  &
    \st{} & \mathcal{M}^{w} \leq M^{wgt},
\end{aligned}
\end{equation}
\begin{equation} \label{eq:bit_allocation_opt-2}
\begin{aligned}
    \underset{\textbf{b}^a \in \mathbb{B}^n}{\mathrm{min}} & \sum_{i=1}^n  \mathcal{D}^a_i   &
    \st{} & \mathcal{M}^{a} \leq M^{act},
\end{aligned}
\end{equation}
where $M^{wgt}$ and $M^{act}$ are memory budgets for weights and activation, respectively, and $M^{wgt} + M^{act} \leq M$. To solve problems \eqref{eq:bit_allocation_opt-1} and \eqref{eq:bit_allocation_opt-2}, we apply the  Lagrangian method proposed in~\cite{Shoham1988EfficientBA}.


To find feasible pairs of $M^{wgt}$ and $M^{act}$, we do a two-dimensional grid search on $M^{wgt}$ and $M^{act}$. The candidates of $M^{wgt}$ and $M^{act}$ are given by uniformly assigning bit-widths $\textbf{b}^w,\textbf{b}^a$ in $\mathbb{B}$. Then, the maximum number of feasible pairs is $|\mathbb{B}|^2$. Thus, we significantly reduce the search space from $|\mathbb{B}|^{2n}$ to at most $2|\mathbb{B}|^{n+2}$ by decoupling \eqref{eq:bit_allocation_opt-0} to \eqref{eq:bit_allocation_opt-1} and \eqref{eq:bit_allocation_opt-2}.

We summarize steps above in Algorithm~\ref{alg}. In the proposed algorithm, we obtain a list $\mathbb{S}$ of potential solutions $(\textbf{b}^w, \textbf{b}^a, n)$ for problem \eqref{eq:optimization}. Then, a solution which minimizes the latency and satisfies the accuracy drop constraint can be selected from the list.

Before ending this section, we make a remark about Algorithm~\ref{alg}.
\begin{remark}
Due to the nature of the discrete nonconvex and nonlinear optimization problem, it is not possible to precisely characterize the optimal solution of \eqref{eq:optimization}. However, our algorithm guarantees $\mathcal{L}(\textbf{b}^w, \textbf{b}^a, n) \leq \min\bigg(\mathcal{L}(\emptyset, \emptyset,0), \mathcal{L}(\textbf{b}^w_e, \textbf{b}^a_e, N)\bigg)$, where $(\emptyset, \emptyset,0)$ is the \CLOUDONLY{} solution, and $ (\textbf{b}^w_e, \textbf{b}^a_e, N)$ is the \EDGEONLY{} solution when it is possible.
\end{remark}

\DontPrintSemicolon
\begin{algorithm}

\SetInd{0.5em}{0.5em}
\SetAlgoLined
\SetAlgoNoEnd

    \KwResult{Split \& bit-widths for weights and activations }
    
    
    $\mathbb{S} \xleftarrow[]{}$ [$(\emptyset, \emptyset, 0)$]. \hspace{5pt} \tcp{\small Initialize with \CLOUDONLY{} solution.} 
    
    $\mathbb{B} \xleftarrow[]{}$ Bit-widths supported by the edge device\\
    

    \For {k in [1, \ldots, $|\mathbb{B}|$] }{
        $M^{wgt}_{k} = \sum_{i=1}^n (\textbf{s}_{i}^{w} \times \mathbb{B}[k])$.\\
        $M^{act}_{k} = \underset{i=1,\ldots,n}{\max}{(\textbf{s}_{i}^{a} \times \mathbb{B}[k])}$.\\
    }
    
    $\mathbb{P} \xleftarrow[]{}$ Solve (\ref{eq:sort_solver_outputs}). \\
    \For {n in $\mathbb{P}$}{
        \For {$M^{wgt}$ in [$M^{wgt}_{1}$, \ldots, $M^{wgt}_{|\mathbb{B}|}$]}{
            $\textbf{b}^{w} \xleftarrow[]{}$ Solve (\ref{eq:bit_allocation_opt-1}). \\
            \For {$M^{act}$ in [$M^{act}_{1}$, \ldots, $M^{act}_{|\mathbb{B}|}$]}{
                \If {$M^{wgt} + M^{act} > M$}{
                    Continue.
                }
                $\textbf{b}^{a} \xleftarrow[]{}$ Solve (\ref{eq:bit_allocation_opt-2}). \\
                \If {$(\textbf{b}^w, \textbf{b}^a, n)$ \text{satisfies} \eqref{eq:con-1}  }{
                    $\mathbb{S}$.append$\big((\textbf{b}^w, \textbf{b}^a, n)\big)$.
                }
            }
        }
    }
    Return $\mathbb{S}.$
    

 \caption{Joint DNN Splitting and Bit Assignment}
 \label{alg}
\end{algorithm}



\subsection{Post-Solution Engineering Steps}
After obtaining the solution of \eqref{eq:optimization}, the remaining work for deployment is to pack the split-layer activations, transmit them to cloud, unpack in the cloud, and feed to the cloud DNN. There are some engineering details involved in these steps, such as how to pack less than 8-bits data types, transmission protocols, etc. We provide such details in the appendix.

%% file: 05_experiments.tex
\section{Experiments} \label{sec:experiment}
In this section we present our experimental results.
\S{\ref{sec:methodology}}-\S\ref{QDMP+Q} cover our simulation-based experiments. We compare \WORK{} with existing state-of-the-art distributed frameworks. We also perform ablation studies to show step by step, how \WORK{} arrives at good solutions compared to existing approaches. In addition, in \S{\ref{sec:lpr}} we demonstrate the usage of \WORK{} for a real-life example of license plate recognition on a low power edge device.



\subsection{Experiments Protocols}
\label{sec:methodology}


Previous studies~\cite{netadapt, chamnet, pnas} have shown that tracking multiply accumulate operations (MACs) or GFLOPs does not directly correspond to measuring latency. We measure edge and cloud device latency on a cycle-accurate simulator based on SCALE-SIM \cite{samajdar2018scale} from ARM software (also used by \cite{choi2020prema, yang2018energy, zhe2020towards2bit}). For edge device, we simulate Eyeriss \cite{eyeriss} and for cloud device, we simulate Tensor Processing Units (TPU). The hardware configurations for Eyeriss and TPU are taken from SCALE-SIM (see Table \ref{tab:acc}). In our simulations, lower bit precision (sub 8-bit) does not speed up the MAC operation itself, since, the existing hardware have fixed INT-8 MAC units. However, lower bit precision speeds up data movement across offchip and onchip memory, which in turn results in an overall speedup. Moreover, we build on quantization techniques \cite{banner2018aciq, nahshan2019loss} using \cite{distiller}. Appendix includes other engineering details used in our setup. It also includes an ablation study on the network bandwidth.

\begin{table}
    \centering
    \caption{Hardware platforms for the simulator experiments}
    \label{tab:acc}
    \renewcommand{\arraystretch}{0.90}%
    \begin{tabular}{C{3cm}C{1.8cm}C{1.8cm}}
    \hline
        Attribute & Eyeriss \cite{eyeriss} & TPU \\ \hline
        On-chip memory & 192 KB & 28 MB \\
        Off-chip memory & 4 GB & 16 GB \\ 
        Bandwidth & 1 GB/sec & 13 GB/sec \\
        Performance & 34 GOPs & 96 TOPs \\ \hline
        Uplink rate & \multicolumn{2}{c}{3 Mbps} \\ 

    \end{tabular}
\end{table}





\subsection{Accuracy vs Latency Trade-off}
\label{sec:tradeoff}

As seen in Sections \ref{sec:formulation} and \ref{sec:solution}, \WORK{} generates a list of candidate feasible solutions that satisfy the memory constraint \eqref{eq:con-1} and error threshold \eqref{eq:con-2} provided by the user. These solutions vary in terms of their error threshold and therefore the down-stream task accuracy. In general, good solutions are the ones with low latency and high accuracy. However, the trade-off between the two allows for a flexible selection from the feasible solutions, depending on the needs of a user. The general rule of thumb is to choose a solution with the lowest latency, which has at worst only a negligible drop in accuracy (i.e., the error threshold constraint is implicitly applied through the task accuracy). However, if the application allows for a larger drop in accuracy, that may correspond to a solution with even lower latency. It is worth noting that this flexibility is unique to \WORK{} and the other methods such as Neurosurgeon ~\cite{neurosurgeon}, QDMP~\cite{QDMP}, U8 (uniform 8-bit quantization), and CLOUD16 (\CLOUDONLY{}) only provide one single solution. 

\begin{figure*}
\centering
\begin{minipage}{.5\textwidth}
  \centering
  \includegraphics[width=0.97\linewidth]{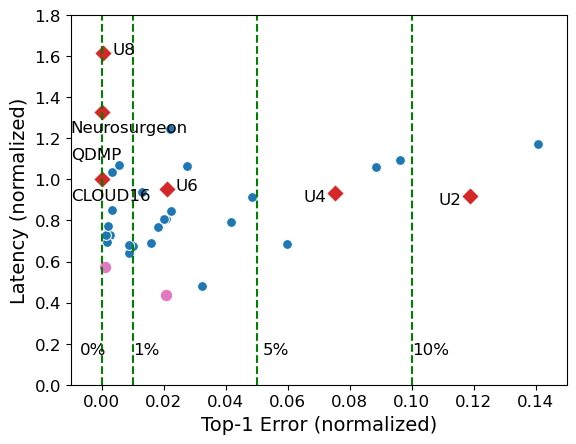}
\end{minipage}%
\begin{minipage}{.5\textwidth}
  \centering
  \includegraphics[width=0.99\linewidth]{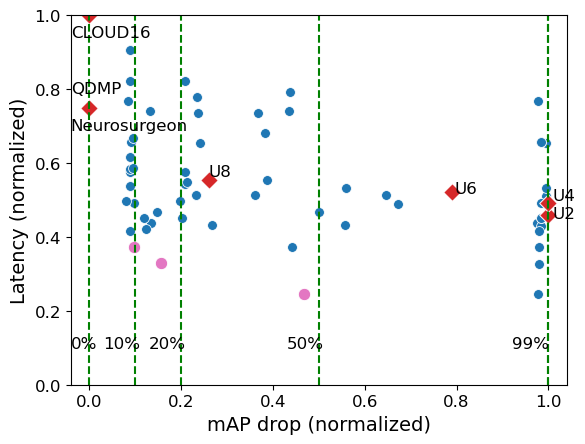}
\end{minipage}
\caption{Accuracy vs latency trade-off for ResNet-50 (left) and Yolo-v3 (right); Towards origin is better. U2, U4, U6 and U8 indicate uniform quantization (\EDGEONLY{}). CLOUD16 indicates FP16 configuration (\CLOUDONLY{}). The green lines show error thresholds which a user can set. \WORK{} can provide different solutions based on different error thresholds (blue and pink). The pink markers show suggested solutions per error threshold.}
\label{plot:resnet50_tradeoff}
\end{figure*}

Fig. \ref{plot:resnet50_tradeoff} shows a scatter plot of feasible solutions for ResNet-50 and Yolov3 DNNs. For ResNet-50, the X-axis shows the Top-1 ImageNet error normalized to \CLOUDONLY{} solution and for Yolo-v3, the X-axis shows the mAP drop normalized to \CLOUDONLY{} solution. The Y-axis shows the end-to-end latency, normalized to \CLOUDONLY{}. The Design points closer to the origin are better. 

\WORK{} can select several solutions (i.e., pink dots in Fig. \ref{plot:resnet50_tradeoff}-left) based on the user error threshold as a percentage of full precision accuracy drop. For instance, in case of ResNet-50, \WORK{} can produce \SPLIT{} solutions with end-to-end latency of 100\%, 57\%, 43\% and 43\%, if the user provides an error threshold of 0\%, 1\%, 5\% and 10\%. For an error threshold of 0\%, \WORK{} selects \CLOUDONLY{} as the solution whereas in case of an error threshold of 5\% and 10\%, \WORK{} selects a same solution.

The mAP drop is higher in the object detection tasks. It can be seen in Fig. \ref{plot:resnet50_tradeoff} that uniform quantization of 2-bit (U2), 4-bit(U4), and 6-bit (U6) result in an mAP drop of more than 80\%. For a user error threshold of 0\%, 10\%, 20\%, and 50\%, \WORK{} selects a solution with an end-to-end latency of 100\%, 37\%, 32\%, and 24\% respectively. For each of the error thresholds, \WORK{} provides a different split point (i.e., pink dots in Fig. \ref{plot:resnet50_tradeoff}-right).  For Yolo-v3, QDMP and Neurosurgeon result in the same split point which leads to 75\% end-to-end latency compared to the \CLOUDONLY{} solution.



\begin{figure*}
 \setarray{dnn-seq}{\textcolor{violet}{\textbf{\CLOUDONLY{}}}, \textcolor{violet}{\textbf{\EDGEONLY{}}}, \textcolor{violet}{\textbf{\SPLIT{}}}}
\begin{tikzpicture}
        \node (dnn) [anchor=south west, inner sep=0pt] {
\includegraphics[width=1\textwidth]{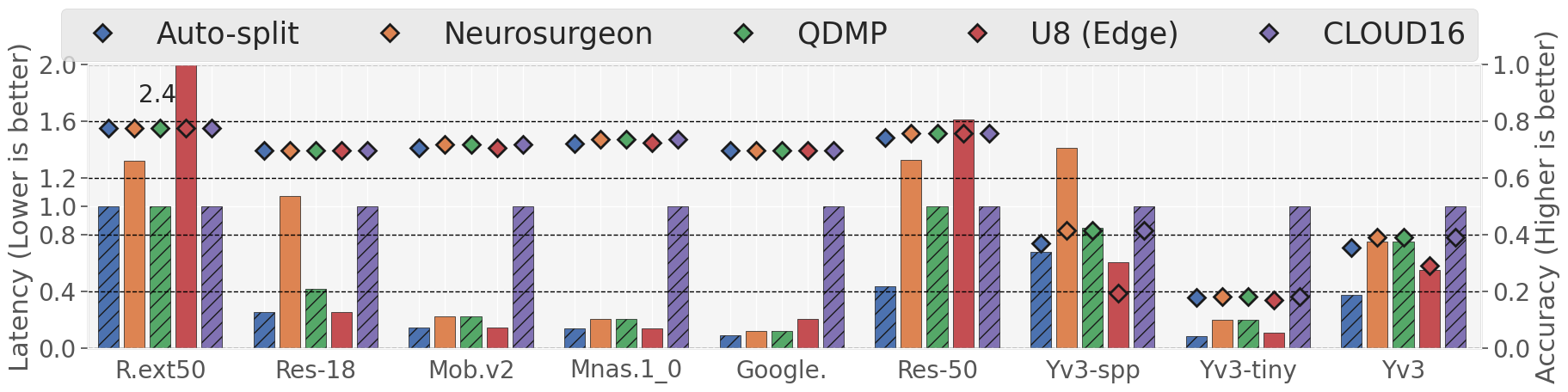}};
\newcounter{cnt}\setcounter{cnt}{1}
\begin{scope}[x={(dnn.south east)},y={(dnn.north west)}]
          \foreach \i/\j in {{(0.157, 0.9)/(0.157,-0.1)},{(0.45,0.9)/(0.45,-0.1)}}
           { \draw [violet, thick, dashed] \i -- \j;
        }
        \foreach \i/\j in {{(0.1,-0.03)/(0.1,0)},{(0.28,-0.03)/(0.28,0)},{(0.6,-0.03)/(0.6,0)}}
           { 
               \node at \i {
       \listarray{dnn-seq}{\thecnt}        };
           \stepcounter{cnt};
        }
        
        \end{scope}
      \end{tikzpicture}

    \caption{Latency (bars) vs Accuracy (points) comparison. Depending on the device constraints, DNN architecture, and network latency, the optimal solution can be achieved from \CLOUDONLY{}, \EDGEONLY{}, or \SPLIT{}. The user error threshold for image classification workloads is 5\%, and for object detection workloads is 10\%.} 
 \label{plot:classification-compare-all}
\end{figure*}
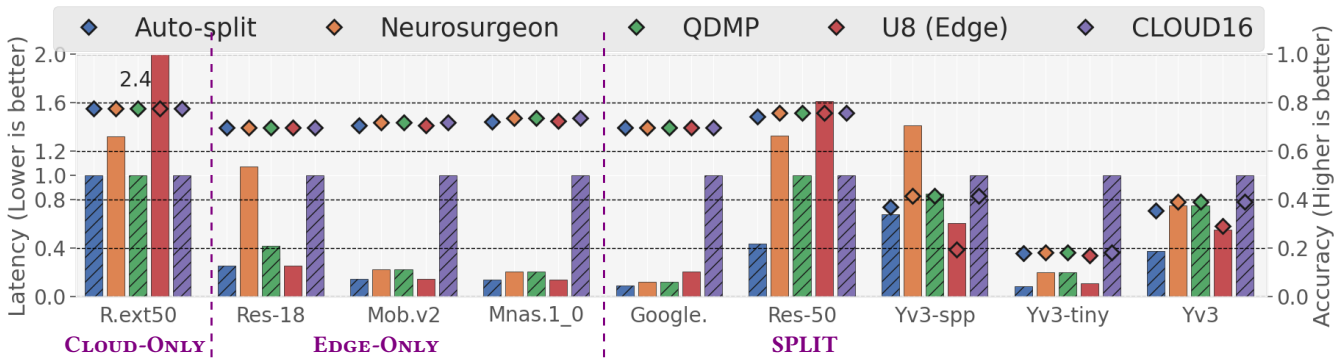

\subsection{Overall Benchmark Comparisons}
\label{sec:overall_comparison}
Fig. \ref{plot:classification-compare-all} shows the latency comparison among various classification and object detection benchmarks. We compare \WORK{}, with baselines: Neurosurgeon ~\cite{neurosurgeon}, QDMP~\cite{QDMP}, U8 (uniform 8-bit quantization), and CLOUD16 (\CLOUDONLY{}). It is worth noting that for optimized execution graphs (see \S \ref{sec:background_distributed}), DADS\cite{dads} and QDMP\cite{QDMP} generate a same split solution, and thus we do not report DADS separately here (see Section 5.2 in \cite{QDMP}).
The left axis in Fig. \ref{plot:classification-compare-all} shows the latency normalized to \CLOUDONLY{} solution which are represented by bars in the plot. Lower bars are better. The right axis shows top-1 ImageNet accuracy for classification benchmarks and mAP (IoU=0.50:0.95) from COCO 2017 benchmark for YOLO-based detection models.

\paragraph{\textbf{Accuracy:}}
Since Neurosurgeon and QDMP run in full precision, their accuracies are the same as the \CLOUDONLY{} solution.

For image classification, we selected a user error threshold of 5\%. However, \WORK{} solutions are always within 0-3.5\% of top-1 accuracy of the \CLOUDONLY{} solution. 
\WORK{} selected \CLOUDONLY{} as the solution for ResNext50\_32x4d, \EDGEONLY{} solutions with mixed precision for ResNet-18, Mobilenet\_v2, and Mnasnet1\_0. For ResNet-50 and GoogleNet \WORK{} selected \SPLIT{} solution.

For object detection, the user error threshold is set to 10\% . Unlike \WORK{}, uniform 8-bit quantization can lose significant mAP, i.e. between 10--50\%, compared to \CLOUDONLY{}. This is due to the fact that object detection models are generally more sensitive to quantization, especially if bit-widths are assigned uniformly.



\paragraph{\textbf{Latency:}} \WORK{} solutions can be: a) \CLOUDONLY{}, b) \EDGEONLY{}, and c) \SPLIT{}. As observed in Fig. \ref{plot:classification-compare-all}, \WORK{} results in a \CLOUDONLY{} solution for ResNext50\_32x4d. \EDGEONLY{} solutions are reached for ResNet-18,  MobileNet-v2, and Mnasnet1\_0, and for rest of the benchmarks \WORK{} results in a \SPLIT{} solution. 
When \WORK{} does not suggest a \CLOUDONLY{} solution, it reduces latency between 32--92\% compared to \CLOUDONLY{} solutions. 

Neurosurgeon cannot handle DAGs. Therefore, we assume a topological sorted DNN as input for Neurosurgeon, similar to \cite{dads,QDMP}. The optimal split point is missed even for float models due to information loss in topological sorting. As a result, compared to Neurosurgeon, \WORK{} reduces latency between 24--92\%.

DADS/QDMP can find optimal splits for float models. They are faster than Neurosurgeon by 35\%. However, they do not explore the new search space opened up after quantizing edge DNNs. Compared to DADS/QDMP, \WORK{} reduces latency between 20--80\%.
Note that QDMP needs to save the entire model on the edge device which may not be feasible. We define QDMP$_E$, a QDMP baseline that saves only the edge part of the DNN on the edge device.


To sum up, \WORK{} can automatically select between \CLOUDONLY{}, \EDGEONLY{},  and \SPLIT{} solutions. For \EDGEONLY{} and \SPLIT{}, our method suggests solutions with mixed precision bit-widths for the edge DNN layers. The results show that \WORK{} is faster than uniform 8-bit quantized DNN (U8) by 25\%, QDMP by 40\%, Neurosurgeon by 47\%, and \CLOUDONLY{} by 70\%.



\subsection{Comparison with QDMP + Quantization}
\label{QDMP+Q}
As mentioned, QDMP (and others) operates in floating point precision. In this subsection, we show that it will not be sufficient, even if the edge part of the QDMP solutions are quantized. To this end, we define QDMP$_E+U_4$, a baseline of adding uniform 4-bit precision to QDMP$_E$. 
Table \ref{tab:dads_vs_auto_split} shows split index and model size for \WORK{}, QDMP$_E$, and QDMP$_E+U_4$.
Note that, ZeroQ reported ResNet-50 with size of 18.7MB for 6-bit activations and mixed precision weights. Also, note that 4-bit quantization is the best that any post training quantization can achieve on a DNN and we discount the accuracy loss on the solution. In spite of discounting accuracay, \WORK{} reduces the edge DNN size by 14.7$\times$ compared to QDMP$_E$ and 3.1$\times$ compared to QDMP$_E$+U4.

\begin{table}
    \centering
    \caption{Comparing QDMP$_E$, \WORK{}, and QDMP$_E$+U4}
    \label{tab:dads_vs_auto_split}
    \renewcommand{\arraystretch}{0.85}%
    \aboverulesep = 0.1pt \belowrulesep = 0.2pt
    \begin{tabular}{C{1.27
    cm}C{1.05cm}C{0.55cm}C{1.05cm}C{0.55cm}C{1.65cm}}

\hline
 & \multicolumn{2}{c}{\WORK{}}  &\multicolumn{2}{c}{QDMP$_E$} & \begin{tabular}[c]{@{}l@{}}QDMP$_E$+U4\end{tabular} \\
 \cmidrule(l{3pt}r{3pt}){2-3}  \cmidrule(l{3pt}r{3pt}){4-5} \cmidrule(l{3pt}r{3pt}){6-6} 
 & Split idx & MB & Split idx & MB & MB \\ \hline
Google. & 18 &  0.4  & 18 & 3.5 & 0.44 \\  
Resnet-50 & 12 & 0.9  & 53 & 50 & 12.5 \\ 
Yv3-spp & 1 & 1.3 & 52 & 99  & 22.2 \\ 
Yv3-tiny & 7 & 8.8  & 7 & 18.2 & 4.44  \\ 
Yv3 & 33 & 13.3  & 58 & 118 & 29.6 \\ 
\end{tabular}
\end{table}

\begin{figure}
    \centering
    \includegraphics[width=1.0\linewidth]{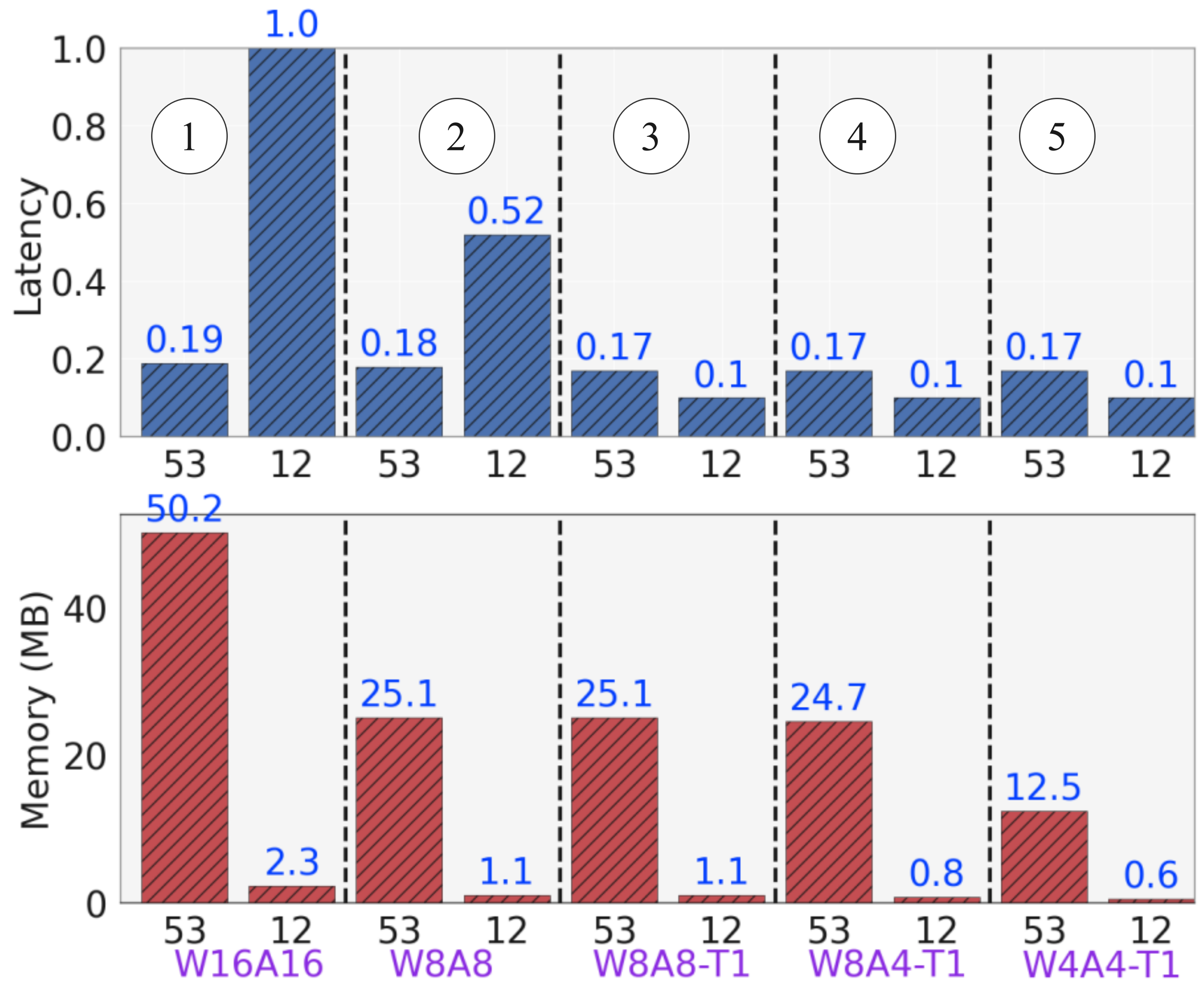}

    \caption{ResNet-50 latency \& memory for \textbf{\WORK{} (split @12}) and \textbf{QDMP (split @53}). W8A8-T1 implies weight, activation, and transmission bit-widths of 8, 8, and 1, respectively. Transmission cost at split 53 is $\simeq 3\times$ less than split 12.}
    \label{plot:latency_dads}

\end{figure}

Next we demonstrate how QDMP$_E$ can select a different split index compared to \WORK{}. Fig. \ref{plot:latency_dads} shows the latency and model size of ResNet-50. \whitecircle{1} : W16A16-T16: Weights, activations, and transmission activations are all 16 bits precision. The transmission cost at split index=53 (suggested by QDMP$_E$) is $\simeq 3\times$ less compared to that of split index=12 (suggested by \WORK{}). Overall latency of split index=53 is 81\% less compared to split index=12.  Similar trend can be seen when the bit-widths are reduced to 8-bit (\whitecircle{2}: W8A8-T8). However, in \whitecircle{3}: W8A8-T1 we reduce the transmission bit-width to 1-bit (lowest possible transmission latency for any split). We notice that split index=12 is 7\% faster compared to split index=53.  Since, \WORK{} also considers edge memory constraints in partitioning the DNN, it can be noticed that the model size of the edge DNN in split index=12  is always orders of magnitude less compared to the split index=53. In \whitecircle{4} and \whitecircle{5} the model size is further reduced. \textit{Overall \WORK{} solutions are $7\%$ faster and $20\times$ smaller in edge DNN model size compared to a QDMP + mixed precision algorithm, when both models have the same bit-width for edge DNN weights, activations, and transmitted activations.}

\subsection{Case Study}
\label{sec:lpr}
We demonstrate the efficacy of the proposed method using a real application of license plate recognition used for authorized entry (deployed to customer sites). In appendix, we also provide a demo on another case study for the task of person and face detection.

Consider a camera as an edge device mounted at a parking lot  which authorizes the entry of certain vehicles based on the license plate  registration. 
Inputs to this system are camera frames and outputs are the recognized license plates (as strings of characters). \WORK{} is applied offline to determine the split point assuming an 8-bit uniform quantization. \WORK{} ensures that the edge DNN fits on the device and high accuracy is maintained. The edge DNN is then passed to TensorFlow-Lite for quantization and is stored on the camera. When the camera is online, the output activations of the edge DNN are transmitted to the cloud for further processing. The edge DNN runs parts of a custom YOLOv3 model. The cloud DNN consists of the rest of this YOLO model, as well as a LSTM model for character recognition.

The edge device (Hi3516E V200 SoC with Arm Cortex A7 CPU, 512MB Onchip, and 1GB Offchip memory) can only run TensorFlow-Lite (cross-compiled for it) with a C++ interface. Thus, we built an application in C++ to execute the edge part, and the rest is executed through a Python interface. 



\begin{table}
    \centering
    \caption{Evaluation of License Plate Recognition solutions}
    \setlength{\tabcolsep}{0.02em}
    \renewcommand{\arraystretch}{0.90}%
    \begin{tabular}{C{3.3cm}C{1.5cm}C{1.8cm}C{1.5cm}} \hline
        Model & Accuracy & Latency & Edge Size \\ \hline
        Float (on edge) & 88.2\% & Doesn't fit & 295 MB\\ 
        Float (to cloud) & 88.2\% & 970 ms & 0 MB \\
        TQ (8 bit) & 88.4\% & 2840 ms & 44 MB \\
        \WORK{} & 88.3\% & \textbf{630 ms} & \textbf{15 MB} \\ 
        \WORK{}(large LSTM) & \textbf{94\%} & \textbf{650 ms} & \textbf{15 MB} \\
    \end{tabular}
    \label{tab:lpr_acc_v1}
\end{table}

Table \ref{tab:lpr_acc_v1} shows the performance over an internal proprietary license plate dataset. The \WORK{} solution has a similar accuracy to others but has lower latency. Furthremore, as the LSTM in \WORK{} runs on the cloud, we can use a larger LSTM for recognition. This improves the accuracy further with negligible latency increase.

%% file: 06_conclusion.tex
\section{Conclusion} \label{sec:conclusion}

This paper investigates the feasibility of distributing the DNN inference between edge and cloud while simultaneously applying mixed precision quantization on the edge partition of the DNN. 
We propose to formulate the problem as an optimization in which the goal is to identify the split and the bit-width assignment for weights and activations, such that the overall latency is reduced without sacrificing the accuracy. This approach has some advantages over existing strategies such as being secure, deterministic, and flexible in architecture. The proposed method provides  a range of options in the accuracy-latency trade-off which can be selected based on the target application requirements.


\balance

%% file: 07_section_supplementary.tex
\section{Appendix} \label{supplementary}
\appendix


\section{Engineering details}

\paragraph{\textbf{Guidelines proposed for users of our system:}} a) For models with float size of < ~50MB, \EDGEONLY{} is likely the optimal solution (on typical edge chips). b) Compared to input image, the activation volumes are generally large for initial layers, but small for deep layers. When a large number of initial layers receive high activation volumes > input image volume, \CLOUDONLY{} is likely the optimal solution. c) For deep but thin networks or when input is high resolution (say >= 416), SPLIT solution is likely optimal.


\paragraph{\textbf{Activation transmission protocol:}}
In practice, we found that python's xmlRPC protocol was orders of magnitude slower compared to using socket programming. The reason is xmlRPC cannot transfer binary data over network, so activations are encoded and decoded into ASCII characters. This adds an extra overhead. Thus, we used socket programming (in C++) for data transmission. Table~\ref{tab:rpc_vs_socket} shows RPC vs socket transmission for a Yolov3 based face detection model. On a single server (31 Gbps), Auto-split takes 1.13s on RPC and 0.27 ms with socket programming. Note that in both xmlRPC or socket programming, quantized activations (say 4-bits) are still stored as ``int8" data type (by padding with zeros). Thus, it requires some pre-processing before transmission for existing edge/cloud devices. Table ~\ref{tab:transmission_api} shows the API of the transmission protocol.

\paragraph{\textbf{Handling sub 8-bit activations for transmission:}} 
\WORK{} may provide solutions with activation layers of lower than 8-bits, e.g. 4-bits. 
To minimize the transmission cost one needs to: 1) either implement ``int4" data type (or lower) for both edge and cloud devices, or 2) pack two 4-bit (or lower) activations into ``int8" data type on the edge device, transmit over the network, and unpack into ``float" data type on the cloud device. We implemented custom $4-bit$ data type to realize low end-to-end latencies. 
For existing devices which do not support $<8-bit$ data type, we also implemented an API to pack/unpack to 8-bit data types. We tried to pack/unpack activations along i) ``Height-Width" and ii) ``Channel" dimensions.
If the edge device supports python, then it is more efficient to use numpy libraries for packing multiple channels of activation ($<8-bit$) to a single 8-bit channel. Table~\ref{tab:packing_overhead} shows details of Height-Width (HW) vs Channel (C) packing \& unpacking overhead of 4-bit activations before transmission (Intel(R) Xeon(R) CPU E5-2690 v4 @ 2.60GHz). For edge devices with C++ interface, SIMD units should be utilized to speed up the packing of activations.

\section{Additional Ablations Studies}

\paragraph{\textbf{Compression of SPLIT layer features for edge-cloud splitting:}}

With the availability of mature data compression techniques, it is natural to consider applying data compression before transmission to cloud, as an attempt to transmit less data, to lower the latency. For \CLOUDONLY{} solutions that means applying image compression (e.g. JPEG), and for \WORK{} corresponds to applying feature compression. Note that this kind of compression depends highly on the edge device support, both in terms of hardware and software, and thus is not always available. Therefore, we study it as an ablation, rather than the default in the algorithm. 

For the \CLOUDONLY{} solutions, we studied the use of JPEG compression, as a candidate with relatively low computation overhead and good compression ratio.

\begin{table}[H]
\centering
\caption{Comparison of RPC vs Socket programming}
\renewcommand{\arraystretch}{0.90}%
\begin{tabular}{C{1.9cm}C{2.3cm}C{0.7cm}C{1.9cm}} 
\hline
\textbf{Benchmark} & \textbf{Img/Act shape} & \textbf{KB} & \textbf{RPC/Socket} \\ \hline 
Cloud-Only & 432,768,3 & 972 & 3566 \\ 
Auto-Split & 36,64,256 & 288 & 3981 \\ 
\end{tabular}
\label{tab:rpc_vs_socket}
\end{table}

\begin{table}[H]
\centering
\caption{API for activation transmission}
\renewcommand{\arraystretch}{0.90}%
\begin{tabular}{C{2.5cm}L{4cm}} 
\hline
\textbf{Data Type} & \textbf{Parameters} \\ \hline
Bytes (int8) & Transmitted Activation \\ 
Float32 & Scale \\ 
Float32 & Zero-point \\ 
List(int32) & Input image shape \\ 
Int8 & \#Bits used for activations \\ 
\end{tabular}
\label{tab:transmission_api}
\end{table}

\begin{table}[H]
\centering
\caption{Packing \& unpacking overhead of 4-bit activations}
\renewcommand{\arraystretch}{0.90}%
\begin{tabular}{C{1.5cm}C{1.4cm}C{0.45cm}C{1.78cm}C{1.4cm}} \hline
\textbf{\small Benchmark} & \textbf{Act shape} & \textbf{KB} & \textbf{\small Height-Width} & \textbf{Channel} \\ \hline
Auto-Split & 36,64,256 & 288 & 1.45 (s) & 0.01 (s) \\ 
\end{tabular}
\label{tab:packing_overhead}
\end{table}

For the task of object detection on Yolov3, at $416\times416$ resolution, and PIL library for JPEG execution, we observed considerable improvements in the solutions, as shown in Table \ref{tab:compression}. However, strong compression results in severe loss of accuracy in the \CLOUDONLY{} case. For feature compression in \WORK{} we followed a similar approach and used the JPEG compression (We initially tried Huffman coding, but the overhead of compression was high and didn't give good solutions). Results of feature compression showed that \WORK{} benefits more on compression ratio, because activations are sparse (~20+\%) and are represented by lower bits e.g. 2bits compared to 8bits (0-255) for input images. For activations we split channels into groups of three and applied JPEG compression.

Note that the overhead of latency for compression/decompression will add another variable to the equation and may result in different splits. For JPEG compression on RaspberryPi3, we measured 28ms overhead for \CLOUDONLY{} and 9ms for \WORK{}. Moreover, compression is better to be lossless or light, otherwise it will damage the accuracy (even if it's not clearly visible). See Fig7-10 in \cite{michaelis2019benchmarking} on object detection or Table1 in \cite{hendrycks2019benchmarking} on classification. At a similar mAP, \WORK{} shows 47\% speed-up compared to \CLOUDONLY{}-QF60. This gap is smaller compared to the non-compressed case of 58\% in Fig. \ref{plot:classification-compare-all}  (Gap depends on architecture/task).

\paragraph{\textbf{Ablation study on network speed:}}
Table \ref{tab:network_speeds} shows results on YOLOv3 (416x416 resolution, and latency is normalized in each case to \CLOUDONLY{}). It is observed from Table \ref{tab:network_speeds} that \WORK{} mAP drops at 20Mbps. The same experiment for YOLOv3SPP (different architecture) resulted in a much better solution. So even at 20Mbps, depending on architecture, there may be good SPLIT solutions.

\paragraph{\textbf{Splitting object detection models:}}
We studied two styles of object detection models: a) Yolo based: Yolov3-tiny, Yolov3, and Yolov3-spp, and b) FasterRCNN with ResNet-50 backbone. Running \WORK{} resulted in SPLIT solutions for Yolo models, but suggested the \CLOUDONLY{} solution for FasterRCNN. In this subsection, we explain the caveats of selecting the backbone network when searching for a \SPLIT{} solution in object detection models. 
For a \SPLIT{} solution to be feasible, the transmission cost of activations at the split point should be at least less than the input image, otherwise a \CLOUDONLY{} solution may have lower end-to-end latency.

\begin{figure}[H]
    \centering
    \includegraphics[width=0.9\linewidth]{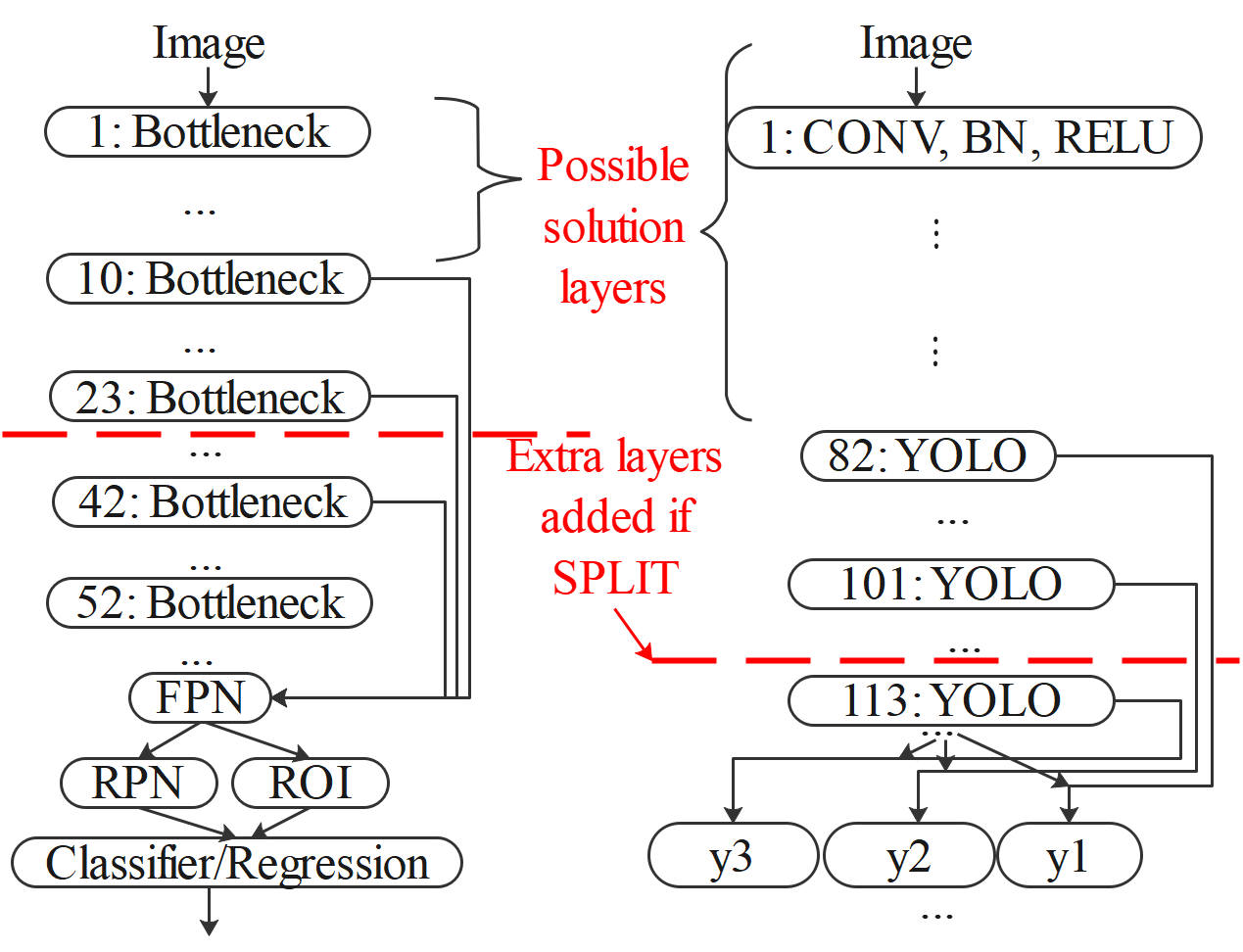}
    \caption{Split layers for FasterRCNN (left) and Yolov3 (right)}
    \label{fig:fasterrcnn_vs_yolo}
\end{figure}

Most Object detection models have a backbone network which branches off to detection and recognition necks/heads. These detection and recognition heads typically collect intermediate features from the backbone network.  For example, Table \ref{tab:intermediate_layers} shows the layer indices of the intermediate layers from which the output features are collected in the Feature pyramid network (FPN) for Faster RCNN and YOLO layers for Yolov3. 
In FasterRCNN, the FPN starts fetching intermediate features from as early as layer index=10, thus in case of a \SPLIT{} solution these features are also required to be transmitted to the cloud, unless entire model is executed in the edge device. As we go deeper in the DNN in search of a \SPLIT{} solution more and more extra intermediate layers need to be transmitted along with the output activation at the split point (see Figure \ref{fig:fasterrcnn_vs_yolo}). Thus, \WORK{} suggests a \CLOUDONLY{} solution for Faster RCNN.

On the contrary, in YOLO based models there are enough number of layers to search for a \SPLIT{} solution before the YOLO layers. For example, in YOLOv3-spp, the search space for a \SPLIT{} solution lies between layer index 0 to 82 and \WORK{} suggested layer index=33 as the split point. 
Therefore, when co-designing an object detection model for distributed inference, it is crucial to start collecting intermediate features as late as possible in the DNN layers. It will be worth looking at a modified Faster RCNN model retrained with FPN network collecting features from layer index=[23,42,52] and see if the accuracy does not drop too much. Such modified model will generate \SPLIT{} solutions.

\paragraph{\textbf{Selecting split points with equal activation volume:}}

Table \ref{tab:split_points} shows the potential split points towards the end of a pretrained ResNet-50 network. After the graph optimizations are applied these layers do not have any other dependencies. The output feature map (OFM) volume is the activation volume for potential split points and the volume difference is the difference between OFM volume and input image volume. For a \SPLIT{} to be valid one primary condition is that the volume difference should be negative.

The end-to-end latency has three components a) edge latency, b) transmission latency and c) cloud latency. In Table \ref{tab:split_points}, layers 46, 49, and 52 have the same shape and thus, same transmission volume difference. Without quantization, layer 46 will be the \SPLIT{} solution, since cloud device is faster than the edge device, and in case of same transmission cost it is preferable to select early layers.

With quantization however, the transmission cost will vary depending on the bit-widths assigned to each layer. The assignment of bit-widths depends on the sensitivity of each layer to compress without losing accuracy. If layer 49 can be compressed more than layer 46, then layer 49 can be selected as a \SPLIT{} solution. The sensitivity of each layer to bit compression can be measured in different ways. \WORK{} considers quantization error of the layers. Other techniques include: Hessian of the feature vector \cite{hawq} or layer sensitivity by quantizing only one layer at a time \cite{zeroq}.

\section{Demo \& Code}
A video demonstration of \WORK{} for the task of person and face detection is available at the following link (best viewed in high resolution). This link also contains a proof-of-concept code for \WORK{} and code to generate the face/person detection demo:
{\noindent \smaller \url{https://drive.google.com/drive/folders/1DX8tS1KeFA2QfPdlzHvaf1JCgnfDOFCN}}

\begin{table} [H]
\centering
\caption{Effect of input or feature compression on solutions.}
\setlength\tabcolsep{1pt} 
\renewcommand{\arraystretch}{0.90}%
\begin{tabular}{C{1.69cm}C{2.15cm}C{1.9cm}C{0.6cm}C{1.75cm}}
\hline
\textbf{Method} & \textbf{JPEG quality factor} & \textbf{Compression ratio} & \textbf{mAP} & \textbf{Normalized latency} \\ \hline
\CLOUDONLY{} & NoCompression & 1$\times$ & 0.39 & 1.0 \\ 
\CLOUDONLY{} & LossLess & 2$\times$ & 0.39 & 0.56 \\ 
\CLOUDONLY{} & 80 & 5$\times$ & 0.38 & 0.23 \\ 
\CLOUDONLY{} & 60 & 8$\times$ & 0.35 & 0.15 \\ 
\CLOUDONLY{} & 40 & 10$\times$ & 0.29 & 0.13 \\ 
\CLOUDONLY{} & 20 & 17$\times$ & 0.22 & 0.09 \\ 
\WORK{}& LossLess & 15$\times$ & 0.35 & 0.08 \\ 
\end{tabular}
\label{tab:compression}
\end{table}

\begin{table} [H]
\centering
\caption{Ablation study on network bandwidth.}
\setlength\tabcolsep{1pt} 
\renewcommand{\arraystretch}{0.90}%
\begin{tabular}{C{1.47cm}C{1.5cm}C{3.5cm}C{1.6cm}}
\hline
\textbf{Model} & \textbf{\small Network bandwidth} & \textbf{Accuracy (mAP) \small \WORK{}/\CLOUDONLY{}} & \textbf{\small Normalized latency}\\ \hline
Yolov3 & 1Mbps & 0.37/0.39 & 0.26/1 \\ 
Yolov3 & 3Mbps & 0.37/0.39 & 0.37/1 \\ 
Yolov3 & 10Mbps & 0.34/0.39 & 0.83/1 \\ 
Yolov3 & 20Mbps & 0.25/0.39 & 0.75/1 \\ 
Yolov3-SPP & 20Mbps & 0.37/0.41 & 0.71/1 \\ 
\end{tabular}
\label{tab:network_speeds}
\end{table}

\begin{table}[H]
\centering
\caption{Layer indices of the collected output activations}
\renewcommand{\arraystretch}{0.90}%
\begin{tabular}{C{2.5cm}C{4.5cm}}
\hline
\textbf{Models} & \textbf{Intermediate Layer indices} \\ \hline
Yolov3-tiny & {{[}16, 23{]}} \\
Yolov3 & 
{{[}82, 94, 106{]}} \\
{Yolov3-spp} & 
{{[}89, 101, 113{]}} \\
{Faster RCNN} &
{{[}10, 23, 42, 52{]}} \\ 
\end{tabular}
\label{tab:intermediate_layers}
\end{table}

\begin{table}[H]
\centering
\caption{Potential splits towards the end of ResNet-50}
\renewcommand{\arraystretch}{0.90}%
\begin{tabular}{C{1cm}C{1.9cm}C{1.15cm}C{1.4cm}C{1.2cm}}
\hline
\textbf{Index} & \textbf{Layer name} & \textbf{Volume} & \textbf{Shape} & \textbf{Vol. Diff} \\ \hline
46 & layer4.0.conv3 & 100,352 & (2048,7,7) & -5076 \\ 
49 & layer4.1.conv3 & 100,352 & (2048,7,7) & -5076 \\ 
52 & layer4.2.conv3 & 100,352 & (2048,7,7) & -5076 \\ 
53 & fc & 1,000 & (1,1000) & -149528 \\ 
-1 & i/p image & 150,528 & (3,224,224) & 0 \\ 
\end{tabular}
\label{tab:split_points}
\end{table}